\documentclass{llncs}
\usepackage{epsfig}
\usepackage{algorithmic}
\usepackage{algorithm}
\usepackage{graphicx}
\usepackage{tabularx} 
\usepackage{booktabs} 
\usepackage{amsmath}
\usepackage{amssymb}
\usepackage{multirow,booktabs}
\usepackage{color}
\usepackage{cite}
\usepackage{multirow}
\usepackage{ulem}
\usepackage{indentfirst}
\usepackage{url}

\begin{document}

\title{ICSI 2014 Competition on Single Objective Optimization \\ (ICSI-2014-BS)}

\author{Ying Tan, Junzhi Li, Zhongyang Zheng}

\institute{Key Laboratory of Machine Perception (Ministry of Education)\\ Department of Machine Intelligence\\ School of Electronics Engineering and Computer Science\\
Peking University, Beijing, P.R. China\\
\email{icsi2014competition@gmail.com}}

\maketitle

\begin{abstract}
 This is the introduction and instruction to the ICSI 2014 Competition on Single Objective Optimization.\\

\textbf{Keywords:} Single Objective Optimization, Swarm Intelligence
\end{abstract}

\section{Introduction}

This competition will focus on single objective optimization, because it is the key and fundamental problem in the Swarm Intelligence. In this competition, we hope to provide a chance for every swarm intelligence algorithm to show its performance and to learn from each other. We welcome any swarm intelligence algorithm to participate in the competition, such as Particle Swarm Optimization, Ant Colony Optimization, Artificial Bee Colony Algorithm, Bat Algorithm, Intelligent Water Drops, Fireworks Algorithm, etc.

The codes for the competition is available at:\\
\indent \indent \indent \indent \url{http://www.ic-si.org/competition/file.zip}

If you have any suggestion, please inform us without hesitation.

\section{Definition}

The task is to minimize the evaluation function:

\[\mathop {\min }\limits_{x \in {{[ - 100,100]}^D}} f(x)\]

There are 30 functions in this competition, all shifted and rotated, which is named as ICSI-2014-Benchmark Suite, i.e., \textbf{ICSI-2014-BS}, for short, and certainly they should be treated as black box problems.

\subsection{Basic Functions}

The following 23 functions are the same in definition as \cite{yao1999evolutionary}, \cite{liang2013problem} and \cite{liang2014problem}.\\

1.Bent Cigar Function

\[f_{1}(x) = x_1^2 + {10^6}\sum\limits_{i = 2}^D {x_i^2} \]

2.High Conditioned Elliptic Function

\[f_2(x) = \sum\limits_{i = 1}^D {{{({{10}^6})}^{\frac{{i - 1}}{{D - 1}}}}x_i^2} \]

3.Neumaire 3 Function

\[x = {D^2}x/100\]

\[f_3(x) = \sum\limits_{i = 1}^D {{{({x_i} - 1)}^2} + \sum\limits_{i = 1}^D {{x_i}{x_{i - 1}} + \frac{{D(D + 1)(D - 1)}}{6}} } \]

4.Discus Function

\[f_4(x) = {10^6}x_i^2 + \sum\limits_{i = 2}^D {x_i^2} \]

5.Different Powers Function

\[f_5(x) = \sqrt {\sum\limits_{i = 1}^D {|{x_i}{|^{2 + 4\frac{{i - 1}}{{D - 1}}}}} } \]

6.Rosenbrock's Function

\[x = 30x/100\]

\[f_6(x) = \sum\limits_{i = 1}^{D - 1} {(100{{(x_i^2 - {x_{i + 1}})}^2} + {{({x_i} - 1)}^2})} \]

7.Alpine Function

\[x = 10x/100\]

\[f_7(x) = \sum\limits_{i = 1}^D {|{x_i}\sin ({x_i}) + 0.1{x_i}|} \]

8.Ackley's Function

\[f_8(x) =  - 20\exp( - 0.2\sqrt {\frac{1}{D}\sum\limits_{i = 1}^D {x_i^2} } ) - \exp(\frac{1}{D}\sum\limits_{i = 1}^D {\cos (2\pi {x_i})} ) + 20 + e\]

9.Weierstrass Function

\[x = x/100\]

\[f_9(x) = \sum\limits_{i = 1}^D {(\sum\limits_{k = 0}^{20} {[{{0.5}^k}\cos(2\pi  \cdot {3^k}({x_i} + 0.5))]} ) - D\sum\limits_{k = 0}^{20} {[{{0.5}^k}\cos(2\pi  \cdot {3^k} \cdot 0.5)]} } \]

10.Griewank's Function

\[x = 600x/100\]

\[f_{10}(x) = \sum\limits_{i = 1}^D {\frac{{x_i^2}}{{4000}} - \prod\limits_{i = 1}^D {\cos (\frac{{{x_i}}}{{\sqrt i }}) + 1} } \]

11.Rastrigin's Function

\[x = 5.12x/100\]

\[f_{11}(x) = \sum\limits_{i = 1}^D {(x_i^2 - 10\cos(2\pi {x_i}) + 10)} \]

12.Katsuura Function

\[x = 5x/100\]

\[f_{12}(x) = \frac{{10}}{{{D^2}}}\prod\limits_{i = 1}^D {{{(1 + i\sum\limits_{j = 1}^{32} {\frac{{|{2^j}{x_i} - \left\lfloor {{2^j}{x_i}} \right\rfloor |}}{{{2^j}}}} )}^{\frac{{10}}{{{D^{1.2}}}}}} - \frac{{10}}{{{D^2}}}} \]

13.Expanded Scaffer's F6 Function

\[g(x,y) = 0.5 + \frac{{({\sin^2}(\sqrt {{x^2} + {y^2}} ) - 0.5)}}{{{{(1 + 0.001({x^2} + {y^2}))}^2}}}\]

\[f_{13}(x) = \sum\limits_{i = 1}^{D - 1} {g({x_i},{x_{i + 1}}) + g({x_D},{x_1})} \]

14.HappyCat Function

\[f_{14}(x) = |\sum\limits_{i = 1}^D {x_i^2 - D} {|^{\frac{1}{4}}} + (0.5\sum\limits_{i = 1}^D {x_i^2 + \sum\limits_{i = 1}^D {{x_i}} } )/D + 0.5\]

15.HGBat Function

\[f_{15}(x) = |{(\sum\limits_{i = 1}^D {x_i^2} )^2} - {(\sum\limits_{i = 1}^D {{x_i}} )^2}{|^{\frac{1}{2}}} + (0.5\sum\limits_{i = 1}^D {x_i^2 + \sum\limits_{i = 1}^D {{x_i}} } )/D + 0.5\]

16.Schwefel's Problem 2.22

\[x = 10x/100\]

\[f_{16}(x) = \sum\limits_{i = 1}^D {|{x_i}| + \prod\limits_{i = 1}^D {|{x_i}|} } \]

17.Schwefel's Problem 1.2

\[f_{17}(x) = \sum\limits_{i = 1}^D {{(\sum\limits_{j = 1}^i {{x_j}} )^2}} \]

18.Schwefel's Problem 2.26

\[x = 500x/100\]

\[f_{18}(x) = \sum\limits_{i = 1}^D {({x_i}\sin(\sqrt {|{x_i}|} ))} \]

19.Penalized Function

\[x = 50x/100\]

\[\mu ({x_i},a,k,m) = \left\{ \begin{array}
{r@{\quad \quad}l}
k{({x_i} - a)^m} & \ {x_i} > a\\
0 & \- a \le {x_i} \le a\\
k{( - {x_i} - a)^m} &\ {x_i} <  - a
\end{array} \right.\]

\[\begin{array}{l}
{f_{19}}(x) = 0.1\{ {\sin ^2}(3\pi {x_1}) + \sum\limits_{i = 1}^{D - 1} {{{({x_i} - 1)}^2}[1 + {{\sin }^2}(3\pi {x_{i + 1}})]} \} \\
\indent \indent \indent + {({x_D} - 1)^2}[1 + {\sin ^2}(2\pi {x_D})]\}  + \sum\limits_{i = 1}^D {\mu ({x_i},5,100,4)}
\end{array}\]

20.Schaffer's F7 Function

\[f_{20}(x) = (\frac{1}{{D - 1}}\sum\limits_{i = 1}^{D - 1} {{{(x_i^2 + x_{i + 1}^2)}^{\frac{1}{4}}} + } {(x_i^2 + x_{i + 1}^2)^{\frac{1}{4}}}{\sin ^2}(50{(x_i^2 + x_{i + 1}^2)^{0.1}}))\]

21.Salomon Function

\[f_{21}(x) = 1 - \cos(2\pi \sum\limits_{i = 1}^D {{x_i}} ) + 0.1\sum\limits_{i = 1}^D {{x_i}^2} \]

\subsection{Composition Functions}

The following 7 functions are newly generated composition functions.\\

22.Well Function

\[{f_{22}(x)}=
\left\{
\begin{array}
    {r@{\quad \quad}l}
    \sum\limits_{i = 1}^D {x_i^2} &  \ \max (x) < 20\\
    400*D &  \  otherwise
\end{array}
\right.\]

23. '8'+'13'+'21'

\[f_{23}(x)=f_8(x)+f_{13}(x)*10+f_{21}(x)*1e-2\]

24. '2'+'9'+'15'+'16'

\[f_{24}(x)=f_2(x)*1e-9+f_9(x)*2+f_{15}(x)*1e-1+f_{16}(x)*5e-2\]

25. '3'+'4'+'7'+'18'

\[f_{25}(x)=f_3(x)*0.25+f_4(x)*1e-9+f_7(x)+f_{18}(x)*1e-2\]

26. '5'+'6'+'12'

\[f_{26}(x)=f_5(x)*1e-5+f_6(x)*1e-7+f_{12}(x)*1e-2\]

27. ('10'+'14'+'20')*'18'

\[f_{27}(x) = f_{18}(f_{10}(x),f_{14}(x),f_{20}(x))\]

28. ('19'+'17'+'1')*'9'

\[f_{28}(x) = f_{9}(f_{19}(x),f_{17}(x),f_{1}(x))\]

29. ('3'+'12'+'15')*'8'

\[f_{29}(x) = f_{8}(f_{3}(x),f_{12}(x),f_{15}(x))\]

30. ('6'+'21'+'14')*'13'

\[f_{30}(x) = f_{13}(f_{6}(x),f_{21}(x),f_{14}(x))\]

\section{Experiment}

1. $D=2,10,30,50$, Search space: $[-100,100]^D$, Maximum evaluation times: $D*10000$.

For each function and each $D$, run 51 times independently and record the best fitness found.

Note that error smaller than $2^{-52} \approx 2.22e-16$ (the eps in matlab) is 0.\\

2. Run the following program 5 times and record the MEAN time consumed as $T1$:

for i = 1 : 300000

\indent  \indent  evaluate(9 , rand(30,1)*200-100);

end

Run your algorithm on function 9 and $D = 30$ for 5 times, and record the MEAN time consumed as $T2$.

\section{Format}

The following things should be included in your paper:

1. Description of your algorithm.

2. The parameters used in your experiment.

3. Experimental environment.

4. $T1$, $T2$ and $(T2-T1)/T1$.

5. For each $D=2,D=10,D=30 and D=50$, show a 30*5 table containing the Max, Min, Mean, Median and Standard deviation of fitness of each function.\\

Besides, you also need to submit 4 result files to the organizers: name\_2d.csv, name\_10d.csv, name\_30d.csv and name\_50d.csv(for example:pso\_2d.csv, pso\_10d.csv, pso\_30d.csv and pso\_50d.csv), with each containing a $30*51$ matrix, showing the best fitness found in each function and each run.\\

The algorithms will be ranked according to their fitness value. The ranking and analysis will be published by the organizers later.

\bibliographystyle{splncs}
\bibliography{ICSI}

\end{document}


\title{Comparison and Ranking Results of Algorithms in ICSI 2014 Competition on Single Objective Optimization}

\author{Junzhi Li and Ying Tan}

\institute{Key Laboratory of Machine Perception (Ministry of Education)\\ Department of Machine Intelligence\\ School of Electronics Engineering and Computer Science\\
Peking University, Beijing, P.R. China\\
\email{ljz@pku.edu.cn, ytan@pku.edu.cn}}

\maketitle

\begin{abstract}
In this technical report, the analyses, comparison and ranking results of the algorithms participating in the ICSI 2014 Competition on Single Objective Optimization(ICSI-2014-BS) are presented.\\

\textbf{Keywords:} Single Objective Optimization, Swarm Intelligence
\end{abstract}

\section{Introduction}

The Fifth International Conference on Swarm Intelligence (ICSI 2014) organized a competition session on Single Objective Optimization. A benchmark suite of 30 evaluation functions called ICSI-2014-BS is set up for this competition session. The competition requires participants to submit their algorithms' evaluation results for 2,10,30 and 50 dimensions and their time complexity index.

There are in total 9 papers submitted to this competition session and 6 papers accepted at last. They are listed in Table \ref{table1}.

\begin{table}[htbp]
  \centering
  \caption{Accepted Papers}
    \begin{tabular}{|c|p{11cm}|}
    \hline
    \# & Title \\
    \hline
    \hline
    33    & Evaluating a Hybrid DE and BBO with Self Adaptation on ICSI 2014 Benchmark Problems \\ \hline
    86    & The Multiple Population Co-evolution PSO Algorithm \\ \hline
    92    & Performance of Migrating Birds Optimization Algorithm on Continuous Functions \\ \hline
    131   & Dynamic Search Fireworks Algorithm for Solving ICSI2014 Competition Problems \\ \hline
    178   & Differential Evolution with Sparks Applied to ICSI 2014 Competition Functions \\ \hline
    180   & Applying Enhanced Fireworks Algorithm to ICSI 2014 Benchmark Suite \\
    \hline
    \end{tabular}%
  \label{table1}%
\end{table}%

The abbreviations of the 6 algorithms are respectively: \textbf{HSDB}, \textbf{MPCPSO}, \textbf{MBO}, \textbf{dynFWA}, \textbf{DESP} and \textbf{EFWA}.

\section{Overview of the Results}

\subsection{Time Complexity}
We defined an index to measure the time complexity of the algorithms:

1. Run the following program 5 times and record the MEAN time consumed as $T1$:

\indent for i = 1 : 300000\\
\indent \indent evaluate(9 , rand(30,1)*200-100);\\
\indent end

2. Run the optimization algorithm on function 9 and $D = 30$ for 5 times, and record the
MEAN time consumed as $T2$.

3. And finally calculate the value $(T2-T1)/T1$ as the metric of the time complexity of the algorithm.

The results are shown in Table \ref{table2}. Algorithms with smaller numbers are faster.

\begin{table}[htbp]
  \centering
  \caption{Time Complexity}
    \begin{tabular}{|c|c|c|c|c|c|c|}
    \hline
    \ Algorithm & HSDB    & MPCPSO    & MBO    & dynFWA   & DESP   & EFWA \\
    \hline
    $(T2-T1)/T1$ & 1.386 & 0.320 & 1.448 & 0.123 & 0.588 & 0.241 \\
    \hline
    \end{tabular}%
  \label{table2}%
\end{table}%

The index to some extent reveals the efficiency of the algorithms. However, we should note that such kind of metrics are very sensitive to the implementation details. Especially on MATLAB platform, a proper optimization of the codes would significantly reduce the time cost. In addition, time complexity of the algorithms is not the main concern in this competition, because for most real world optimization problems, the evaluation is overwhelmingly expensive.

\subsection{Mean Fitness Value}
We limit the evaluation times of each function to $10000 \times D$ for each dimensionality $D=2,10,30,50$. Table \ref{table3},\ref{table4},\ref{table5} and \ref{table6} present the mean fitness value of each dimensionality obtained by the algorithms over 51 independent runs. The best fitness values are highlighted.

\begin{table}[H]
  \centering
  \caption{Mean Fitness $(D=2)$}
    \begin{tabular}{ccccccc}
    \toprule
    \ Func\textbackslash Alg & HSDB  & MPCPSO & MBO   & dynFWA & DESP  & EFWA \\
    \midrule
    1     & \textbf{0.00E+00} & 1.75E-02 & 1.25E+03 & 1.56E+02 & 4.99E+02 & 4.32E+02 \\
    2     & \textbf{0.00E+00} & 2.54E+01 & 6.09E+02 & 4.90E+02 & 8.89E+02 & 5.07E+02 \\
    3     & \textbf{1.67E+00} & 1.67E+00 & 1.67E+00 & 1.67E+00 & 1.67E+00 & 1.67E+00 \\
    4     & \textbf{0.00E+00} & 3.95E+00 & 3.76E+02 & 1.79E+02 & 3.23E+02 & 3.21E+02 \\
    5     & \textbf{0.00E+00} & 3.47E-06 & 5.52E-01 & 3.85E-03 & 3.23E+00 & 7.91E-04 \\
    6     & \textbf{0.00E+00} & 5.66E-23 & 1.16E+00 & 1.21E-03 & 1.77E+00 & 2.70E-04 \\
    7     & \textbf{0.00E+00} & 1.59E-06 & 2.55E-03 & 9.70E-04 & 4.62E-03 & 6.15E-03 \\
    8     & \textbf{0.00E+00} & 1.03E-15 & 3.74E+00 & 2.45E-02 & 3.04E-01 & 8.19E-01 \\
    9     & \textbf{-4.00E+00} & -4.00E+00 & 2.73E-01 & -3.98E+00 & -3.75E+00 & -3.99E+00 \\
    10    & \textbf{-9.97E-01} & -9.95E-01 & 2.29E-01 & -9.76E-01 & -7.92E-01 & -9.85E-01 \\
    11    & \textbf{0.00E+00} & 6.20E-15 & 3.05E-01 & 7.02E-05 & 7.29E-01 & 1.63E-05 \\
    12    & 4.89E+01 & 4.95E+01 & \textbf{5.41E-01} & 4.93E+01 & 6.96E+01 & 5.17E+01 \\
    13    & \textbf{6.89E-03} & 1.07E-02 & 2.31E-02 & 1.55E-02 & 7.29E-02 & 2.64E-02 \\
    14    & 4.76E-03 & \textbf{5.66E-04} & 7.23E-01 & 6.47E-02 & 6.49E-02 & 1.61E-01 \\
    15    & 7.53E-04 & \textbf{2.41E-04} & 9.02E-01 & 1.74E-02 & 2.59E-01 & 3.81E-02 \\
    16    & \textbf{0.00E+00} & 6.76E-18 & 7.61E-02 & 2.31E-03 & 1.48E-01 & 5.43E-04 \\
    17    & \textbf{0.00E+00} & \textbf{0.00E+00} & 1.83E-01 & 3.63E-03 & 1.95E+00 & 8.96E-05 \\
    18    & -8.38E+02 & -7.72E+02 & \textbf{-8.39E+02} & -8.38E+02 & -6.49E+02 & -6.25E+02 \\
    19    & \textbf{0.00E+00} & 4.44E-16 & 1.07E-01 & 1.14E-03 & 2.69E-03 & 9.26E-04 \\
    20    & \textbf{0.00E+00} & \textbf{0.00E+00} & 2.79E+00 & \textbf{0.00E+00} & 1.06E-13 & \textbf{0.00E+00} \\
    21    & \textbf{0.00E+00} & \textbf{0.00E+00} & 4.24E-01 & 1.98E-03 & 1.57E-02 & 3.53E-07 \\
    22    & \textbf{0.00E+00} & 5.08E-33 & 3.51E-01 & 2.27E-03 & 9.13E-01 & 5.40E-05 \\
    23    & 8.97E+00 & 8.97E+00 & \textbf{6.74E+00} & 8.99E+00 & 1.05E+01 & 9.16E+00 \\
    24    & \textbf{-4.53E+00} & -4.51E+00 & 5.49E+02 & -4.42E+00 & -3.90E+00 & -4.03E+00 \\
    25    & -1.60E+00 & \textbf{-1.63E+00} & 4.23E+02 & -1.57E+00 & -9.05E-01 & -1.10E+00 \\
    26    & \textbf{5.86E-01} & 6.04E-01 & 2.57E+00 & 6.68E-01 & 8.93E-01 & 7.81E-01 \\
    27    & \textbf{-3.74E+07} & -3.73E+07 & -3.83E+04 & -3.74E+07 & -2.93E+07 & -2.99E+07 \\
    28    & -5.22E+00 & \textbf{-5.74E+00} & -8.85E-01 & -5.24E+00 & -5.32E+00 & -5.07E+00 \\
    29    & 2.00E+01 & 2.00E+01 & \textbf{1.29E+01} & 2.00E+01 & 2.00E+01 & 2.00E+01 \\
    30    & 3.91E-01 & \textbf{3.19E-01} & 3.60E-01 & 3.39E-01 & 8.68E-01 & 6.31E-01 \\
    \bottomrule
    \end{tabular}%
  \label{table3}%
\end{table}%

\begin{table}[H]
  \centering
  \caption{Mean Fitness $(D=10)$}
    \begin{tabular}{ccccccc}
    \toprule
    Func\textbackslash Alg & HSDB  & MPCPSO & MBO   & dynFWA & DESP  & EFWA \\
    \midrule
    1     & \textbf{5.91E+00} & 4.16E+03 & 5.74E+07 & 3.18E+04 & 9.12E+05 & 5.86E+04 \\
    2     & \textbf{2.87E+01} & 3.11E+03 & 8.60E+02 & 9.49E+03 & 1.04E+03 & 7.52E+03 \\
    3     & \textbf{1.70E+02} & 1.70E+02 & 2.44E+02 & 1.70E+02 & 1.71E+02 & 1.70E+02 \\
    4     & \textbf{3.00E-01} & 7.86E+01 & 1.02E+03 & 6.48E+01 & 5.93E+00 & 1.89E+01 \\
    5     & \textbf{9.36E-03} & 2.51E-02 & 1.36E+00 & 6.31E-02 & 2.86E+00 & 1.12E-01 \\
    6     & \textbf{5.02E+00} & 5.37E+00 & 1.90E+03 & 6.56E+00 & 5.83E+01 & 7.81E+00 \\
    7     & \textbf{1.30E-04} & 8.02E-03 & 8.04E-01 & 1.64E-02 & 9.09E-02 & 8.51E-01 \\
    8     & \textbf{1.55E-01} & 2.48E+00 & 9.38E+00 & 2.03E+00 & 4.00E+00 & 1.66E+01 \\
    9     & \textbf{-1.99E+01} & -1.85E+01 & 4.72E+00 & -1.84E+01 & -1.77E+01 & -1.48E+01 \\
    10    & \textbf{-8.35E+00} & -7.91E+00 & 1.69E+00 & -7.24E+00 & -6.80E+00 & -2.16E+00 \\
    11    & \textbf{1.05E+00} & 5.40E+00 & 2.64E+01 & 6.54E+00 & 1.18E+01 & 2.04E+01 \\
    12    & \textbf{4.52E-01} & 4.76E-01 & 1.52E+00 & 4.70E-01 & 4.62E-01 & 4.96E-01 \\
    13    & \textbf{4.00E-01} & 6.23E-01 & 2.15E+00 & 9.80E-01 & 1.16E+00 & 2.11E+00 \\
    14    & \textbf{5.16E-02} & 5.82E-02 & 7.99E+00 & 8.83E-02 & 1.21E-01 & 7.24E-01 \\
    15    & \textbf{4.85E-02} & 1.66E-01 & 1.20E+02 & 1.03E-01 & 2.32E-01 & 3.02E-01 \\
    16    & \textbf{5.50E-03} & 2.16E-01 & 2.66E+00 & 2.67E-01 & 2.29E-01 & 2.45E-01 \\
    17    & \textbf{7.46E-04} & 3.04E-02 & 1.77E+03 & 2.57E-03 & 1.30E-01 & 2.44E-02 \\
    18    & -2.27E+03 & -1.94E+03 & -2.91E+03 & -2.31E+03 & -1.97E+03 & \textbf{-3.00E+03} \\
    19    & \textbf{6.86E-04} & 8.28E-03 & 4.22E+00 & 2.42E-02 & 3.51E-02 & 8.23E-01 \\
    20    & 5.11E-03 & \textbf{1.31E-03} & 1.03E+01 & 3.95E-01 & 2.89E-01 & 9.57E+00 \\
    21    & \textbf{9.99E-02} & 2.94E-01 & 1.48E+01 & 2.88E-01 & 5.98E-01 & 2.29E-01 \\
    22    & \textbf{6.69E-04} & 3.46E-03 & 1.17E+02 & 4.80E-02 & 3.47E+00 & 9.16E-02 \\
    23    & \textbf{1.94E+01} & 2.47E+01 & 3.83E+01 & 2.90E+01 & 3.03E+01 & 3.89E+01 \\
    24    & \textbf{-3.53E+01} & -3.16E+01 & 1.03E+03 & -3.29E+01 & -3.04E+01 & -2.18E+01 \\
    25    & \textbf{4.29E+01} & 4.29E+01 & 1.44E+03 & 4.36E+01 & 4.48E+01 & 4.37E+01 \\
    26    & \textbf{6.71E-03} & 8.40E-03 & 2.33E+03 & 7.92E-03 & 8.95E-03 & 9.84E-03 \\
    27    & \textbf{-1.02E+08} & -2.82E+07 & -1.37E+04 & -7.83E+07 & -5.25E+07 & -9.75E+07 \\
    28    & -5.38E+00 & -5.66E+00 & 1.51E+01 & \textbf{-5.77E+00} & -5.40E+00 & -5.37E+00 \\
    29    & 2.00E+01 & 2.00E+01 & 2.14E+01 & \textbf{2.00E+01} & 2.00E+01 & 2.00E+01 \\
    30    & \textbf{1.01E+00} & 1.02E+00 & 1.25E+00 & 1.02E+00 & 1.04E+00 & 1.04E+00 \\
    \bottomrule
    \end{tabular}%
  \label{table4}%
\end{table}%

\begin{table}[H]
  \centering
  \caption{Mean Fitness $(D=30)$}
    \begin{tabular}{ccccccc}
\toprule
    Func\textbackslash Alg & HSDB  & MPCPSO & MBO   & dynFWA & DESP  & EFWA \\
    \midrule
    1     & 7.64E+04 & \textbf{4.01E+04} & 2.78E+08 & 5.48E+04 & 1.37E+07 & 6.64E+04 \\
    2     & 5.05E+03 & 9.24E+03 & \textbf{7.70E+02} & 1.26E+04 & 1.21E+03 & 1.23E+04 \\
    3     & 4.52E+03 & 4.52E+03 & 2.75E+04 & 4.52E+03 & 4.99E+03 & \textbf{4.52E+03} \\
    4     & 7.46E+00 & 6.72E+01 & 1.03E+03 & 3.03E+01 & \textbf{1.67E+00} & 8.91E+00 \\
    5     & \textbf{3.75E-02} & 4.04E-02 & 1.11E+00 & 7.79E-02 & 5.29E+00 & 7.12E-02 \\
    6     & 2.91E+01 & \textbf{2.91E+01} & 8.60E+03 & 3.64E+01 & 2.04E+02 & 3.03E+01 \\
    7     & \textbf{7.91E-03} & 1.83E-02 & 2.64E+00 & 1.20E-02 & 7.88E-02 & 2.12E+00 \\
    8     & \textbf{9.06E-01} & 3.04E+00 & 1.08E+01 & 2.75E+00 & 4.39E+00 & 1.93E+01 \\
    9     & \textbf{-5.83E+01} & -5.35E+01 & 1.58E+01 & -5.33E+01 & -5.29E+01 & -1.07E+01 \\
    10    & \textbf{-2.48E+01} & -2.39E+01 & 3.77E+00 & -2.22E+01 & -2.05E+01 & -1.27E+01 \\
    11    & \textbf{2.92E-02} & 1.47E+00 & 1.15E+02 & 1.07E+01 & 2.45E+01 & 7.82E+01 \\
    12    & 1.43E-02 & 1.40E-02 & 3.60E+00 & 1.41E-02 & \textbf{1.33E-02} & 1.41E-02 \\
    13    & 3.04E+00 & \textbf{1.91E+00} & 9.16E+00 & 2.98E+00 & 3.53E+00 & 7.32E+00 \\
    14    & 7.32E-02 & \textbf{2.11E-02} & 8.85E+00 & 4.61E-02 & 5.50E-02 & 1.94E+00 \\
    15    & \textbf{9.48E-02} & 4.55E-01 & 2.95E+02 & 2.41E-01 & 3.65E-01 & 8.91E-01 \\
    16    & \textbf{2.57E-01} & 7.66E-01 & 7.50E+00 & 6.31E-01 & 1.62E+00 & 4.92E-01 \\
    17    & 2.02E+00 & 7.11E+00 & 5.47E+04 & 2.17E+00 & \textbf{1.75E-01} & 1.07E+00 \\
    18    & -5.82E+03 & -2.50E+03 & -4.49E+03 & -5.03E+03 & -4.53E+03 & \textbf{-9.01E+03} \\
    19    & \textbf{5.04E-03} & 2.50E-02 & 1.18E+01 & 1.95E-01 & 1.19E-01 & 1.27E+00 \\
    20    & 2.01E-01 & 6.43E-01 & 8.06E+00 & 2.76E+00 & 4.19E+00 & \textbf{1.15E-05} \\
    21    & \textbf{2.14E-01} & 7.23E-01 & 3.07E+01 & 8.87E-01 & 1.41E+00 & 6.86E-01 \\
    22    & 1.00E-01 & \textbf{4.81E-02} & 3.21E+02 & 7.58E-02 & 1.63E+01 & 6.73E-02 \\
    23    & 4.43E+01 & \textbf{3.96E+01} & 1.09E+02 & 4.98E+01 & 5.35E+01 & 9.27E+01 \\
    24    & \textbf{-1.15E+02} & -1.06E+02 & 9.04E+02 & -1.04E+02 & -1.03E+02 & -8.55E+01 \\
    25    & 1.13E+03 & 1.13E+03 & 1.93E+04 & 1.13E+03 & 1.26E+03 & \textbf{1.13E+03} \\
    26    & 4.38E-04 & 4.65E-04 & 8.21E+03 & \textbf{4.23E-04} & 6.31E-04 & 4.31E-04 \\
    27    & -4.78E+08 & -1.09E+08 & -5.16E+03 & -2.64E+08 & -2.45E+08 & \textbf{-6.11E+08} \\
    28    & -5.48E+00 & -5.74E+00 & 5.52E+01 & \textbf{-5.80E+00} & -5.49E+00 & -5.52E+00 \\
    29    & 2.00E+01 & 2.00E+01 & 2.16E+01 & 2.00E+01 & \textbf{2.00E+01} & 2.00E+01 \\
    30    & \textbf{1.02E+00} & 1.04E+00 & 1.42E+00 & 1.06E+00 & 1.08E+00 & 1.09E+00 \\
    \bottomrule
    \end{tabular}%
  \label{table5}%
\end{table}%

\begin{table}[H]
  \centering
  \caption{Mean Fitness $(D=50)$}
    \begin{tabular}{ccccccc}
    \toprule
    Func\textbackslash Alg & HSDB  & MPCPSO & MBO   & dynFWA & DESP  & EFWA \\
    \midrule
    1     & \textbf{5.60E+04} & 1.62E+05 & 3.87E+08 & 6.22E+04 & 4.03E+06 & 9.01E+04 \\
    2     & 9.04E+03 & 7.11E+04 & \textbf{8.31E+02} & 1.33E+04 & 1.06E+06 & 1.42E+04 \\
    3     & 2.09E+04 & 2.10E+04 & 2.69E+05 & \textbf{2.09E+04} & 2.89E+04 & 2.09E+04 \\
    4     & 7.66E+00 & 1.17E+02 & 1.06E+03 & 2.07E+01 & 5.16E+02 & \textbf{5.09E+00} \\
    5     & \textbf{3.54E-02} & 8.53E-02 & 1.05E+00 & 4.99E-02 & 2.84E+00 & 6.16E-02 \\
    6     & 4.88E+01 & 4.99E+01 & 1.00E+04 & \textbf{4.85E+01} & 9.59E+01 & 4.91E+01 \\
    7     & \textbf{2.01E-02} & 4.40E-02 & 3.35E+00 & 2.44E-02 & 8.12E-02 & 4.38E+00 \\
    8     & \textbf{1.52E+00} & 2.96E+00 & 9.89E+00 & 3.23E+00 & 3.08E+00 & 2.01E+01 \\
    9     & \textbf{-9.61E+01} & -9.09E+01 & 2.42E+01 & -8.78E+01 & -9.05E+01 & -1.29E+01 \\
    10    & \textbf{-4.35E+01} & -3.12E+01 & 4.62E+00 & -3.82E+01 & -1.50E+01 & -2.78E+01 \\
    11    & \textbf{3.61E-02} & 9.70E-02 & 1.61E+02 & 7.04E-01 & 2.21E+00 & 1.14E+02 \\
    12    & \textbf{2.76E-03} & 2.91E-03 & 3.72E+00 & 2.91E-03 & 3.04E-03 & 2.83E-03 \\
    13    & \textbf{4.09E+00} & 7.37E+00 & 1.69E+01 & 4.15E+00 & 1.45E+01 & 1.53E+01 \\
    14    & 1.69E-02 & \textbf{1.08E-02} & 8.04E+00 & 2.87E-02 & 1.11E-01 & 2.00E+00 \\
    15    & \textbf{4.26E-01} & 4.26E-01 & 3.69E+02 & 4.58E-01 & 4.60E-01 & 1.74E+00 \\
    16    & \textbf{5.89E-01} & 1.31E+00 & 1.06E+01 & 9.37E-01 & 1.60E+00 & 7.08E-01 \\
    17    & 6.43E+00 & 6.10E+01 & 1.60E+05 & 5.98E+00 & 3.42E+02 & \textbf{3.41E+00} \\
    18    & -7.44E+03 & -3.22E+03 & -5.89E+03 & -4.84E+03 & -8.94E+03 & \textbf{-1.50E+04} \\
    19    & \textbf{9.37E-03} & 3.29E-01 & 1.46E+01 & 2.11E-01 & 7.94E-01 & 8.86E-01 \\
    20    & 3.11E-01 & 9.90E+01 & 6.76E+00 & 8.93E+00 & 2.87E+03 & \textbf{4.39E-06} \\
    21    & \textbf{4.06E-01} & 9.14E-01 & 3.79E+01 & 1.23E+00 & 1.76E+00 & 9.37E-01 \\
    22    & \textbf{3.73E-02} & 1.03E-01 & 4.07E+02 & 4.98E-02 & 3.78E+00 & 7.45E-02 \\
    23    & \textbf{4.43E+01} & 8.57E+01 & 1.86E+02 & 5.76E+01 & 1.61E+02 & 1.48E+02 \\
    24    & \textbf{-1.89E+02} & -1.79E+02 & 8.56E+02 & -1.73E+02 & -1.79E+02 & -1.50E+02 \\
    25    & \textbf{5.23E+03} & 5.26E+03 & 1.18E+05 & 5.23E+03 & 7.31E+03 & 5.23E+03 \\
    26    & 2.29E-04 & 2.34E-04 & 9.69E+03 & 2.27E-04 & 2.31E-04 & \textbf{2.25E-04} \\
    27    & -8.43E+08 & -2.62E+08 & -4.00E+04 & -3.60E+08 & \textbf{-2.39E+09} & -2.32E+09 \\
    28    & -5.54E+00 & \textbf{-5.85E+00} & 9.52E+01 & -5.78E+00 & -5.56E+00 & -5.54E+00 \\
    29    & 2.00E+01 & \textbf{2.00E+01} & 2.17E+01 & 2.00E+01 & 2.00E+01 & 2.00E+01 \\
    30    & 1.04E+00 & \textbf{1.04E+00} & 1.45E+00 & 1.11E+00 & 1.09E+00 & 1.14E+00 \\
    \bottomrule
    \end{tabular}%
  \label{table6}%
\end{table}%

\section{Ranking Results}
In order to rank the performance of the algorithms, we conducted a round robin on them. For each function, if algorithm $\bf{A}$ performs significantly better (examined by a t-test with 95\% confidence level) than algorithm $\bf{B}$, then $\bf{A}$ gets one point. The point numbers are presented in Table \ref{table7}.

\begin{table}[H]
  \centering
  \caption{Point Number}
    \begin{tabular}{|c|c|c|c|c|c|c|}
    \hline
    Dim\textbackslash Alg & HSDB  & MPCPSO & MBO   & dynFWA & DESP  & EFWA \\
    \hline
    \hline
    2     & 113   & 104   & 23    & 60    & 16    & 42 \\ \hline
    10    & 127   & 88    & 9     & 82    & 46    & 41 \\ \hline
    30    & 104   & 85    & 8     & 75    & 63    & 66 \\ \hline
    50    & 117   & 77    & 12    & 91    & 45    & 73 \\ \hline
    Sum   & 461   & 354   & 52    & 308   & 170   & 222 \\
    \hline
    \end{tabular}%
  \label{table7}%
\end{table}%

\bf{So, the final ranking is:}

\begin{enumerate}
\item Hybrid DE and BBO with Self Adaptation

\item Multiple Population Co-evolution PSO Algorithm

\item Dynamic Search Fireworks Algorithm

\item Enhanced Fireworks Algorithm

\item Differential Evolution with Sparks

\item Migrating Birds Optimization Algorithm
\end{enumerate}

\rm{}
Congratulations to YuJun Zheng and XiaoBei Wu, whose paper \itshape{Evaluating a Hybrid DE and BBO with Self
Adaptation on ICSI 2014 Benchmark Problems} \rm{} wins the ICSI 2014 Competition on Single Objective Optimization. HSDB shows a significant advantage over other algorithms on all the dimensionality.

\section{Acknowledgements}
Thank all the authors for their participation. And we would like to give special thanks to Dr. Jane Jing Liang for her valuable advices.

This work was supported by National Natural Science Foundation of China (NSFC) under Grant No. 61375119, 61170057 and 60875080.

\bibliographystyle{splncs}
\bibliography{ICSI}